\documentclass[3p,authoryear,preprint,review, 11pt]{elsarticle}

\usepackage[utf8x]{inputenc}
\usepackage[T1]{fontenc}
\usepackage{subfigure}

\usepackage{amssymb}
\usepackage{graphicx}
\usepackage[colorinlistoftodos]{todonotes}
\usepackage[colorlinks=true, allcolors=blue]{hyperref}
\usepackage{enumitem}
\usepackage{makecell}
\usepackage{algorithm}
\usepackage{algpseudocode}
\usepackage{booktabs}
\usepackage{url}
\usepackage{verbatim}

\usepackage{multicol}
\usepackage{float}

\usepackage{titlesec}
\titleformat{\paragraph}
{\normalfont\normalsize\itshape}{\theparagraph}{1em}{}
\titlespacing*{\paragraph}
{0pt}{3.25ex plus 1ex minus .2ex}{1.5ex plus .2ex}

\usepackage{lineno}
\usepackage{setspace}
\usepackage{amsmath}
\DeclareMathOperator*{\argmin}{\arg\!\min}

\begin{document}

\begin{frontmatter}
	
	


%
%

%

\title{Digital Twin Applications in Urban Logistics: An Overview}
%
%
%
\author[TUE]{Abdo Abouelrous\fnref{*}}
	\ead{a.g.m.abouelrous@tue.nl}
	\fntext[*]{corresponding author}
	\author[TUE]{Laurens Bliek}
	\ead{l.bliek@tue.nl}
        \author[TUE]{Yingqian Zhang}
	\ead{yqzhang@tue.nl}

%
\address[TUE]{Department of Information Systems, Faculty of Industrial Engineering and Innovation Sciences, Technical University Eindhoven, 5612 AZ, Eindhoven, The Netherlands}
%
%
\begin{abstract}
Urban traffic attributed to commercial and industrial transportation is observed to largely affect living standards in cities due to external effects pertaining to pollution and congestion. In order to counter this, smart cities deploy technological tools to achieve sustainability. Such tools include Digital Twins (DT)s which are virtual replicas of real-life physical systems. Research suggests that DTs can be very beneficial in how they control a physical system by constantly optimizing its performance. The concept has been extensively studied in other technology-driven industries like manufacturing. However, little work has been done with regards to their application in urban logistics. In this paper, we seek to provide a framework by which DTs could be easily adapted to urban logistics networks. To do this, we provide a characterization of key factors in urban logistics for dynamic decision-making. We also survey previous research on DT applications in urban logistics as we found that a holistic overview is lacking. Using this knowledge in combination with the characterization, we produce a conceptual model that describes the ontology, learning capabilities and optimization prowess of an urban logistics digital twin through its quantitative models. We finish off with a discussion on potential research benefits and limitations based on previous research and our practical experience.
\end{abstract}

\end{frontmatter}

\section{Introduction}

Urban Logistics has been growing rapidly in recent years due to rising consumer demand and online shopping, among other trends relating to population growth and urbanization (\cite{savelsbergh201650th}). As a result, operational planning and policy-making in urban logistics has become increasingly complex. The associated challenges require the development of `smart' technologies that can assist with planning and resource allocation in urban logistics (\cite{buyukozkan2021smart}). Such technologies are often attributed with smart cities that enjoy a fortified technological infrastructure by which city data can be collected and processed to improve decision-making by stakeholders within the city.

Smart technologies are normally derivatives of  Artificial Intelligence (AI), which has managed to acquire significant interest from research and industry with its promising capabilities. Specifically, AI has witnessed many industrial applications in urban logistics to deal with real-life planning challenges as discussed in \cite{jucha2021use}, \cite{sonneberg2019autonomous} and \cite{shi2019online}.
Among the various AI-driven technologies presented to tackle urban logistics problems, there is one that we are particularly interested in, namely because of its holistic approach in combining knowledge from different computational models. More precisely, some of the most important AI techniques such as learning and optimization could be easily embedded in its framework. We refer to this as the Digital Twin (DT). 

The DT term has been first proposed in 2003 at Michigan University by Professor Grieves to prescribe product life-cycle management as explained in \cite{semeraro2021digital}. Despite dating back almost two decades, there is still no standardized industrial or academic definition for what constitutes a DT. The term is generally used to refer to any virtual replica of a real-life model but provides no guidelines or technical requirements on the functionalities of this replica. This often leads to confusion among parties with regards to what a DT is and how it is distinct from many existing simulation/decision support systems. We believe that this confusion has presented an obstacle to the development of research in DTs, while this paper aims to overcome this confusion by providing a clear characterization of DTs in urban logistics.

In particular, the definition of what a DT is, its functional requirements and conceptual components were often determined by the application context (\cite{semeraro2021digital}). For the large part, DT applications have largely been manufacturing-based as in \cite{rosen2015importance} and \cite{tao2019digital}. On the other hand, there has been very limited focus on the area of logistics compared to manufacturing as \cite{hasse2019digital} mentions, despite its importance and emphasis on how it could benefit from Big Data analytics associated with DTs \cite{pan2021smart}. This benefit only grows with time as cities are becoming increasingly smarter and collect data from a multitude of sources \cite{white2021digital}. Specifically, by using this data, DTs could be used to improve the quality of life, mobility and services of the inhabitants of a city \cite{botin2022digital}.  

In order to define a framework for building DTs, we first have to provide a characterization of urban logistics operations in terms of the key factors that govern it. The human aspect plays an important role in the urban environment as determined by the stakeholders and their interactions (\cite{lagorio2017urban}). This aspect is less significant in other domains such as manufacturing where robotic equipment operating in an exclusive environment is mainly responsible for decision-making and implementation which guarantees some consistency in input and output. Urban environments, on the other hand, are seen to be more complex (\cite{rydin2012shaping}). As a result, a specific characterization upon which urban logistics DTs could be built is needed.

In response to the aforementioned problems, we strive to deliver the following contributions through this study:
 \begin{enumerate}[noitemsep]
    \item Characterize urban logistics operations in terms of defining factors for dynamic decison-making. We arrive at four major input components we take to be resources, stakeholders, KPIs and measures.  
     \item Summarize previous findings from literature on DTs in urban logistics in terms of definition, technical anatomy, functionalities and set-up methodology. To the best of our knowledge, there is no existing research that contains a holistic overview of all three topics as previous papers tend to focus on a subset of these topics.
      \item Provide a framework on the conceptual anatomy of urban logistics DTs in terms of the AI methods employed. This refers to a three-part framework on knowledge base, machine learning and 
      optimization.
     \item Specify potential opportunities and challenges in future research of urban logistics DTs based on previous literature and our practical experience.
 \end{enumerate}
   
 That said, the rest of this article is organized as follows.
 Section \ref{Urban Logistics} provides the characterization of urban logistics (Contribution 1). Section \ref{DT: UL} discusses previous literature on DTs in urban logistics (Contribution 2) in terms of definition, technical requirements \& capabilities and set-up methodology. Section \ref{conceptual model} proposes the conceptual framework (Contribution 3). Section \ref{Adv & Lim} lists the potential benefits and limitations (Contribution 4).

\section{Urban Logistics}
\label{Urban Logistics}

We first start by providing a definition for DTs in urban logistics. To arrive at a domain-specific definition - as conventional in the research of DTs - we first need to establish the important aspects of an urban logistics environment that a DT ought to cover. The authors of \cite{savelsbergh201650th} defines urban logistics as the efficient and effective transportation of goods in urban regions. The scope of our research thus reduces to transportation problems only in contrast to other logistical problems that deal with warehousing, shift-scheduling etc.

To construct a DT model for urban logistics, it is imperative to identify key factors that characterize this spectrum of operations. \cite{anand2012genclon} provide an ontology for urban logistics whereby these factors are identified. We only consider a subset of the factors that we find relevant as input for dynamic decision-making in urban logistics. These are stakeholders, Key Performance Indicators (KPI)s, resources and measures. Note that decisions are also a key factor in this ontology. However, they are an output factor in response to the aforementioned input factors. We elaborate on the four input factors below.

\subsection{Stakeholders} 
The government, businesses and citizens in the urban logistics supply chain are referred to as stakeholders. For a detailed survey on the roles of stakeholders in urban logistics, we refer to \cite{lagorio2016research}. The authors of \cite{lagorio2017urban} explain how stakeholders have an integral function in defining the ecosystem of urban logistics network through their interests, interactions and decisions. On the other hand, 

\subsection{KPIs}

As explained above, stakeholders have interests. These interests translate to objectives which are measured using KPIs as stated in \cite{morana2015sustainable}. KPIs could be used to guide optimization procedures for logistical operations as they can be used to represent objective functions. An example can be found in \cite{van2020evaluating} who use an agent-based simulation to verify routing schemes. The schemes are assessed by predetermined KPIs representing the objective functions. Sustainable urban logistics networks KPIs normally include - and are not restricted to -  CO2 emissions, cost, lead time and delivery travel time as given in \cite{sarraj2014interconnected}. For an exhaustive list of some of the most popular KPIs for urban logistics, we refer to \cite{griffis2007aligning} and \cite{gunasekaran2007performance}. 

\subsection{Resources}
Resources refer to all the available resources possessed by all stakeholders in the urban logistics network. \cite{szmelter2020assessing} explains how urban logistics resources fall into four categories, namely material, human, capital and information. Material resources include machines like trucks, IT platforms etc. Human resources refer to all the laborers involved in executing decisions in the urban logistics supply chain and the decision-makers themselves. Capital refers to the financial resources. Information refers to the intellectual resources such as knowledge and experience. 

\subsection{Measures}
Lastly, the measures represent the rules and regulations under which the resources of the digital twins operate. Most often these measures include regulations imposed by policy-makers such as in \cite{russo2010classification} and \cite{munuzuri2005solutions} and include restricting goods vehicle access to certain roads such as in heavily congested residential areas. Note that these rules correspond to constraints that are not embedded in the resources themselves, unlike the maximum capacity of a vehicle for example, but rather imposed by the rule of law. 

Being practical constraints, the measures could be used to configure modelling constraints when setting up mathematical optimization models for urban operations. Resulting solutions ought to respect the constraints, so that the corresponding real-life decisions remain feasible. Conversely, quantitative approaches could also be adopted by policy-makers to study the effects of their proposed measures as explained in \cite{cardenas2017city}.

\section{Digital Twins}
\label{DT: UL}

We provide a general definition for DTs based on a survey of definitions from several applications from \cite{semeraro2021digital}. We then inspect literature that specifically discussed DT applications in urban logistics and use this to assemble information on the technical anatomy, functionalities and set-up of an urban logistics DT.Other literature with different application scopes is disregarded here. 

\subsection{Definition}
\label{DT definition}

The abundant definition as per previous research initiatives goes along the lines of DTs being digital reconstructions of real-life physical systems that mimic the behavior of the systems and their integral components through real-time linkage with the systems.

So far, the definition above resembles closely that of simulation in the sense that they are both virtual representations of physical entities. However, the part on real-time linkage bears huge importance in defining the distinction as explained in \cite{semeraro2021digital}. A DT is synched to the physical system - which is a city in our case - in the sense that the status of the twin always corresponds to the current real-life status, and is updated once the real-life status changes. The status of the twin should not depict anything that is not happening in real-time. For instance, traffic jams should not be depicted by the virtual model when traffic is not building up in reality, even when empirical traffic data implies otherwise. Nonetheless, the DT may visualize phenomena that are not currently happening in real-life through its simulation tools as will be explained below. The distinction between the simulated visualizations and the status of the virtual model, however, ought to remain clear.

In particular, \cite{semeraro2021digital} states that “The Digital Twin should evolve synchronously with the real system along its whole life cycle", thereby modifying its initial configuration to adapt to the current situation. That said, the DT does not only update its status but also the quantitative models it embeds. The autonomous procedure by which it does will be stressed below. The targeted outcome of this procedure is that the modelling accuracy of the DT steadily improves.

Returning to the distinction from simulation, the DT's dynamic replication of a physical system resembles emulation more closely as \cite{semeraro2021digital} explains. The DT ensures that the integral components of the system it models are accurately represented through the data it collects about them rather than simply model the general behavior of the system. The increased level of representation offered by emulation serves a considerable benefit as it provides a closer replication of reality in contrast to the more static concept of simulation.

DTs also control the physical system through data that they transfer to it. A unidirectional data flow - from the physical to the virtual model - is not sufficient for a virtual model to be labelled a DT. \cite{semeraro2021digital} state that the interaction with the physical system should be bidirectional as data collected from the physical space updates the virtual model, while the physical twin improves its operational performance by exploiting knowledge acquired from the virtual model's processing of the data. For instance, an analytic insight generated from the DT regarding traffic should be communicated to the logistic planners/traffic controllers to assist them with decision-making. This may not be the case with a simulation model, since it is not linked to the physical system and thus, the analysis it produces need not be relevant for the current situation. 

To conclude their findings, \cite{semeraro2021digital} provide the following definition of a DT:

\begin{center}
“A set of adaptive models that emulate the behavior of a physical system in a virtual system getting real time data to update itself along its life cycle. The digital twin replicates the physical system to predict failures and opportunities for changing, to prescribe real-time actions for optimizing and/or mitigating unexpected events observing and evaluating the operating profile system”.
\end{center}

The definition above is not specific to urban logistics indeed. Nonetheless, we feel that the literature is in pressing need for an established basis upon which other features of DTs could be built for different applications.

\subsection{Technical Anatomy}
\label{Technical Anatomy}

From a software perspective, there are many components in a DT. Generally speaking, the literature refers to components such as Internet of Things (IOT), cloud-computing and Application Programming Interfaces (APIs) such as in \cite{moshood2021digital}. The software engineering aspect, however, is of minor interest to us. Instead, we adopt a conceptual approach explaining how AI capabilities could aid real-time decision making in an urban setting when integrated in the DT framework. For a detailed discussion on how software is integrated into the DT framework, we refer to \cite{botin2022digital}. 

\begin{figure}[ht]
    \centering
    \includegraphics[width=.8\linewidth]{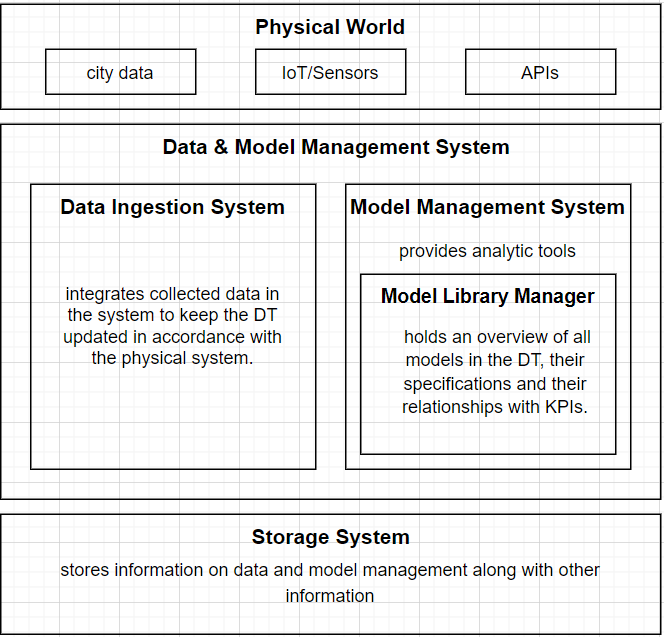}
\caption{General technical anatomy of an urban logistics DT based on \cite{belfadel2021towards}.}
    \label{DT conceptual anatomy}
\end{figure}

From a technical standpoint, \cite{belfadel2021towards} propose a framework into the anatomy of twins. We present an illustration based on their anatomy in Figure \ref{DT conceptual anatomy}. They explain that a typical DT model is composed of the following hierarchies: the top-level hierarchy which is known as the Physical World, the intermediate level known as the Data and Model Management System and the bottom level known as the Storage System. 

The Physical World represents the external physical entities and sensors such as city operational data, IoT entities and sensors, and APIs. This level is linked to the Data and Model Management System that is composed of two main systems: Data Ingestion System (DIS) and Model Management System (MMS). The DIS aims to integrate contextual data entities to the system and keep the DT updated about the status of the physical system. The MMS manages the models library which is a set of software applications that provide analytic tools. Most importantly, the MMS contains the Model Library Manager (MLM) which holds an overview of all available models in the DT, their specifications and their relationships with KPIs. The Storage System stores information on the management of model libraries and managed data along with other general storage as simulation scenarios, their configuration, related models and data etc. 

Returning to the architecture described above, this is the task of the Decision System in the MMS which is responsible for assessing scenarios in terms of how likely they may occur, examines the KPIs of the selected scenario and recommends the necessary interventions in the physical world through the DIS to achieve the predicted outcome. 

\subsection{Functionalities}
\label{functionalities}

The main application area of DTs that is relevant for our study is smart cities. The introduction of DTs is supposed to help overcome many issues that urban models generally suffer from.  \cite{nochta2019governance} provide examples of these issues such as a degree of simplification of urban processes, shortages in data requirements, complications involved in data collection and inadequately handling human behavior and its implications, among others.  

From a computational perspective, \cite{marcucci2020digital} mention that DTs in smart cities “describe, capture and simulate policy (both real and potential) implications of alternative solutions for optimizing them with respect to a given objective or set of them”. \cite{schislyaeva2021innovations} survey functionalities pertaining to DTs of logistic networks on the higher-level, emphasizing on how they can dynamically prescribe and optimize the urban physical system. They also discuss how DT simulation models could be used for stress tests while their predictive analytics tools can be used to predict the state of their physical counterpart. \cite{gutierrez2021data} state that DTs can predict possible future scenarios and evaluate them to find appropriate responses for the most likely ones.

These functionalities are of utter relevance to urban logistics. For instance, Big Data from traffic movements could be used to calculate travel times which can in turn be used to set-up optimization models for vehicle routing and configure simulation models to verify the proposed routing plans. Moreover, the analytics embedded in the twin can be used to predict possible vehicle failures judging by their working circumstances or other internal diagnostics that would otherwise be unobserved. \cite{schislyaeva2021innovations} confirm this by saying that DTs retain data that can not be (easily) obtained from the physical model.

\cite{gutierrez2021data} mention that DTs can raise alerts when exceptional situations are detected so that controllers can intervene. Anomaly detection could be pressing in some situations such as when a vehicle has been observed to remain stationary for a prolonged period of time. This could be the result of an accident in a distant location that would have otherwise remained undetected. Exception handling is especially relevant when physical assets exist in unsafe environments as \cite{moshood2021digital} suggests, which improves the safety protocols. Not to mention the typical operational disruption consequent to accidents that could be efficiently managed by DTs where the difficulty of real-time re-planning is better addressed by real-time data connectivity and great computational power, improving the responsiveness of a Logistics Service Provider (LSP). This is because DTs can not only predict if a problem might happen, but also propose a solution.

DTs also encompass large sets of KPIs that are generated from its diagnostics.  \cite{gutierrez2021data} mentions that DTs can support accurate calculation of performance indicators of logistics operations through their scenario prediction and assessment mechanisms. \cite{moshood2021digital} also mentions how DT sensors could be used to generate new data types, which as \cite{gutierrez2021data} suggests could create even new relevant KPIs. As \cite{belfadel2021towards} explains,“The decisive factor is how this data is processed further in order to offer real added value. In this context, the added value is created with the help of KPIs tailored precisely to the targeted application”. Additionally, KPIs play an even more important role in the learning process of a DT, something that will be extensively discussed below.

There is an additional benefit brought by the visual effects of a DT, and that is supply chain visibility. \cite{moshood2021digital} explain that supply chain visibility depends on an organization's ability to be transparent and clear about its internal and external processes of its supply chain. To that end, organizations have to determine the logistics operations that are most affected by lack of transparency and clarity, and devising techniques by which data could be easily exchanged between all participants. The visibility factor is not only important because of the enhanced interpretability it allows by visually depicting operations, but it also plays a collaborative part due to the involvement of multiple stakeholders along the supply chain who can be easily informed in a standardized manner of expected outcomes to any process. \cite{moshood2021digital} confirm the importance of this by stating that it is essential to provide as much information as possible at the higher-degrees of strategic decision-making that can impact the supply chain as a whole. \cite{marcucci2020digital} also stresses the importance of DTs in translating complex ideas into more intuitive ones through visualization. 

\cite{moshood2021digital} reinforce the technological prowess of DTs by stressing how they could automate monotonous tasks that could be subject to human error, which is a benefit that could only come with its real-time connectivity. The advantages of automated processes are countless and can be explored in manifold applications. Furthermore, \cite{schislyaeva2021innovations} mention that it can troubleshoot remote equipment and perform remote maintenance as an extension of its automation capabilities.

The concept of DTs is normally coined with a self-learning feature where the virtual system consistently tries to improve its modelling of the physical system through the data it retrieves from it. \cite{gutierrez2021data} mention “a learning process based on the KPIs” process, where the modelling parameters of the DT can be calibrated by comparing the actual outcome of the operations with results from the simulation and optimization models. Therefore, DTs can learn from daily operations using machine learning models hat facilitate the acquisition and accumulation of knowledge from the urban environment. In turn, accumulated knowledge can be used to make predictions about the outcomes of future operations when data about these operations is not available.

In continuation of this learning framework, \cite{kalaboukas2021implementation} presents a study about how `cognitive' DTs in agile supply chains increase their knowledge base as they learn more from data obtained from the physical model. They state that DTs should try to relate their predicted outcomes to the actual observed ones by learning how Unpredictable Desired (UD: desirable outcomes that were not predicted) and Unpredictable Undesired (UU: undesirable outcomes that were not predicted) events affect the physical counterpart. By employing this framework, DTs can enhance their learning capabilities over time as it offers a guideline to train and improve DT models through knowledge gathered from past UD and UU events.

To sum up, and as \cite{nochta2019governance} explains, the introduction of DTs compliments the 4th industrial revolution where ``moving from a period of relative data scarcity to an era of ‘digital abundance’ may enable'' the generation of more accurate models based on real-time Big Data of higher quality that can describe urban logistics processes on a greater level of detail than before.

\subsection{Set Up \& Illustration}
\label{Set Up}

In principle, setting up a DT is a complex process.  
\cite{moshood2021digital} states that is a relatively new area of research and that its precise implementations are scarce. This goes in synergy with the findings of \cite{belfadel2021towards}, implying “that existing architectures are too generic for usage in logistics”. For smart cities, there have been several partial implementations such as Cambridge \cite{kalaboukas2021implementation}, Lyon \cite{belfadel2021towards} and cities in the Netherlands as we will explain below. \cite{botin2022digital} also surveys other studies of smart city DTs in Asia and Europe, of which a subset is dedicated to urban logistics. Some research initiatives such as \cite{ivanov2020digital} propose the concept of a DT of a city from a governance perspective, with limited focus on urban logistics that involves other stakeholders such as LSPs. 

As an emphasis on the challenge associated with building DTs, \cite{moshood2021digital} stresses that ``a completely integrated Digital Twins is a long-term approach that does not happen immediately insisting that it will be long before it can be used by industry''. Much of the difficulty is attributed to the intense technological requirements such as Internet of Things Sensors, Cloud computing etc. Consequently, \cite{moshood2021digital} proposes to start simple and focus on maintaining the accuracy of data while incrementally reducing the chance of human error.

\cite{marcucci2020digital} provides an example of a collaborative initiative between policy-makers in the contest of Living-Labs. They suggest that Living-Labs are the most up-to-date data-driven methodology tackling the problem of managing urban logistics. The major idea is to involve all potential stakeholders in the urban logistics network in the design of the DT to agree on common objectives and functionalities. The concept is being tested in cities like Gothenburg, London and Rome. There, Living Labs are developed to create efficient and shared solutions among stakeholders. 

\cite{moshood2021digital} also states that knowledge ought to be exchanged among the multiple stakeholders, a factor catalyzed by visibility. An extensive analysis on sociotechnical interaction among government, industry and consumers is conducted in \cite{nochta2019governance}, who argue that there is a need to look beyond technological factors and incorporate a distinct societal aspect into the design and implementation of DTs if they were to make any resonating changes in the practical modelling of urban environments.

Another important aspect regarding the design of a DT is the modeling one. \cite{marcucci2020digital} also explains that DTs should strive to provide a simplified version of the physical model as they should never replicate the physical system in every detail, as that would not make them models anymore that we could use to efficiently study urban environments. The purpose of the DT as a model is to abstract the complex environment of a city in a limited number of variables. This sets many implications on what factors to include in the model. More variables could be included in the future as the requirements of its user base expand.

For the learning part, \cite{kalaboukas2021implementation} say that it is necessary to create a data set for which observed behaviors of the virtual model can be categorized among UP and UU events. This process, however, involves considerable data-manipulation and cleaning. For instance, they mention how “removing unpredicted and undesirable behaviors” is necessary when configuring a DT model to ensure that the training data is undisturbed. This is subsequent to experimenting with different scenarios and concluding on a particular desired behavioral model, so that only behaviors of interest are included in the design.

There are other less technical factors that should also be carefully contemplated when designing a DT. These could include legal issues regarding data-sharing. Since the DT is a broad model that encompasses multiple stakeholders, some with conflicting objectives, it is important to govern how data is shared and used such that each and every party knows exactly what it needs to know. This is discussed in detail in \cite{kalaboukas2021implementation}.
Working in environments with multiple stakeholders often also requires the introduction of common operational norms, rules and objectives. The DT needs to be aware of those rules that reflect the priorities, policies and other terms of collaboration among stakeholders.

\begin{figure}
    \centering
    \includegraphics[width=.9\linewidth]{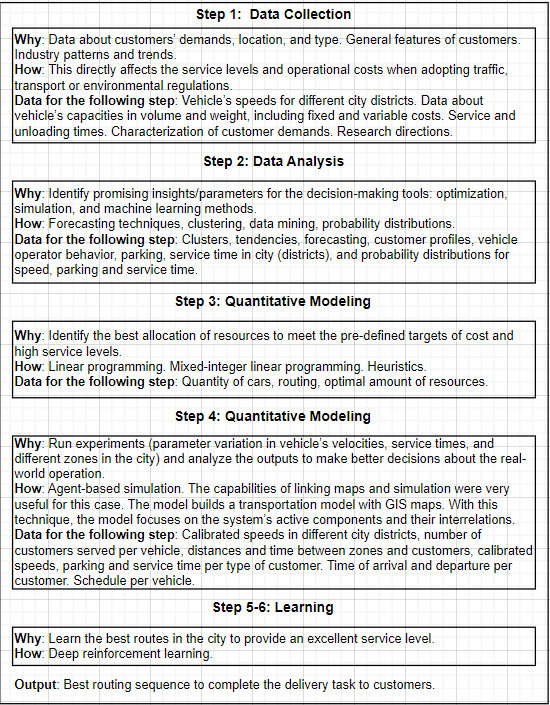}
\caption{DT set up methodology based on \cite{gutierrez2021data}.}
    \label{DT set up}
\end{figure}

All factors considered, \cite{gutierrez2021data} provide a methodology on the steps taken to set-up an urban logistics DT for an LSP. Figure \ref{DT set up} provides an illustration based on their methodology. The 6-step procedure starts with data collection. The data types collected should be dictated by the availability of data and interests of LSP using the DT. In Step 2, and after the data has been collected, it has to be suitably processed for a particular purpose such as devising diagnostic statistics that prescribe the operational context and can then be fed to a mathematical model. For the latter purpose in Step 3, a mathematical optimization model could be set up by which decisions are made. The planning then has to be verified by means of a simulation in Step 4 in a more representative setting where the behavior of the city is better captured than in a simplified optimization model. Once the solution has been verified, it is set to be implemented in Step 5, with KPIs on its actual performance in the city being generated in real-time. The realized KPIs from Step 5 are compared with the estimated ones in Step 4. Major deviations are corrected for  in Step 6 through configured learning processes such as reinforcement learning in order to ensure more accurate modelling and better decision-making in the future.

The design proposed in Figure \ref{DT set up} is specific to an LSP in a developing city. An urban logistics DT in general could be utilized by different stakeholders such as policy-makers and consumers in different cities which may or may more developed in line with the Living Lab approach presented above. This poses new implications on the design of the urban logistics DT. For instance, different data types could be collected in Step 1 depending on the requirements of the stakeholder(s). Additionally, LSPs may deal with different problems for which they have different approaches that deviate from the sequential procedure in \cite{gutierrez2021data}. To that end, different stakeholders may require different `versions' of the DT. This is adhered in \cite{schislyaeva2021innovations} which states that ``One object can have more than one twin, with different models created for different users and use cases". Therefore, it is necessary to come up with an ontology upon which the general design of an urban logistics DT could be established to accommodate for all these possible variations. Consequently, we propose a conceptual model in Section \ref{conceptual model} that discusses such an ontology and the integration of AI tools in the urban logistics DT.

\section{Conceptual Model}
\label{conceptual model}

In this section, we present our own conceptual model of DTs based on AI methodology. Our model is identified by three components, namely a knowledge base, machine learning and mathematical optimization.

\subsection{Knowledge Base}
\label{Knowledge base}

Using the characterization in Section \ref{Urban Logistics} and the set-up methodology of \cite{gutierrez2021data} in Section \ref{Set Up}, we are able to devise an ontology expressed by means of a knowledge graph. This ontology is similar to the one used in \cite{anand2012genclon} in terms of knowledge and relation representation. To specify the high-level entities, we use a mix of the five key factors in Section \ref{Urban Logistics} and define several others based on the set-up in Figure \ref{DT set up}. Specifically, we have the following entities:
stakeholders, resources, KPIs, measures, decisions, data, statistical analysis tools, mathematical optimization, simulation and machine learning where the former five are discussed in Section \ref{Urban Logistics}, while each of the latter five corresponds to a step in Figure \ref{DT set up}.  

We are yet to elaborate on the latter five. We deem the Data entity to be self-explanatory as it represents any data collected by the DT through sensors for example, so we do not expand it further here. Specifically, the other four entities below that use the data represent relevant AI tools that can be used to solve urban logistics related-problems. Statistical analysis of data has been used in many urban logistics studies such as \cite{zou2020evaluation} and \cite{alho2015utilizing}, mathematical optimization in \cite{montoya2017electric} and \cite{dabia2017exact}, simulation modelling in \cite{jlassi2018simulation} and \cite{karakikes2018techniques} and machine learning in \cite{el2020machine} and \cite{giuffrida2022optimization}, to give some examples.

Referring to the technical architecture in Section \ref{Technical Anatomy}, the four AI entities would be embedded in the MLM. The MLM is an integral component of DTs as it embeds the quantitative tools that express its analytical capabilities. In turn, human controllers should consistently strive to improve its content so as to equip the DT sufficiently to analyze very complex processes and bring added value through optimization. In order to arrive at a compact design of the ontology, we merge these four into a single high-level entity which we refer to as the AI component. This is because their relationships with other entities in the ontology and among each other are more or less identical. 

\begin{figure}
\centering
        \includegraphics[width=.9\linewidth]{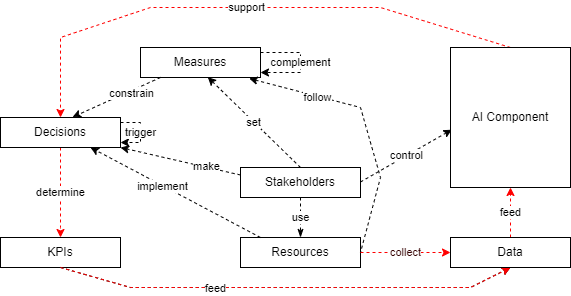}
    \caption{Our proposed high-level urban logistics DT ontology.}
    \label{ontology}
\end{figure}

At the high-level, our ontology is given in Figure \ref{ontology} where all entities are defined as well as their associated relations. The feedback-loop constituted by the bidirectional data exchange to optimize current operations and learn from past ones to improve future ones is given by the red arrows. In particular, data is collected from resources and used as input to the AI component. The AI component processes the data to support decision-making through optimization. Once decisions have been implemented, KPIs are generated which are collected again as data to evaluate the decisions and learn from them.

\begin{figure}
\centering
        \includegraphics[width=.65\linewidth]{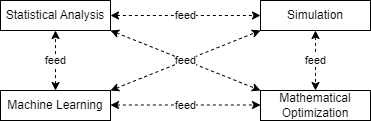}
    \caption{Our proposed ontology for the AI component in Figure \ref{ontology}.}
    \label{AI component}
\end{figure}

At a secondary level, the ontology of the AI component itself is given in Figure \ref{AI component}. The four components interact regularly with each other to support their functionalities. The output of one component can be used as the input to another, in contrast to the ordered pattern introduced in Figure \ref{DT set up}. Figure \ref{AI component} could, thus, be viewed as a generalization of Figure \ref{DT set up}. It can also be viewed as an arbitrary pipeline of models. The same logic holds the other way round as any pipeline of these models could be categorized under the ontology of Figure \ref{AI component}. To demonstrate this, we provide an example below for solving a  Vehicle Routing Problem (VRP) variant. We explain how a pipeline could be constructed from an algorithmic set-up, how it integrates into the framework of the DT and how the DT uses its capabilities to improve the AI models through its real-time connectivity.

However, we first provide a brief overview of previous research on how the aforementioned AI tools complement each other in decision-making for urban logistics - with a focus on routing operations - in the following section. Our focus on routing stems from our observation of the academic interest surrounding problems such as VRP and the importance of solving VRP variants for the daily operations of LSPs.

\subsection{AI Pipeline Construction}

There are some parts of the AI component that we pay little attention to. For instance, we consider statistical analysis to be a more traditional method where the computing powers of AI are less relevant. For simulation, we provide a brief discussion as it remains a relevant tool for digital-twins for scenario-assessment especially. We are most interested in the combination of machine-learning and optimization as dictated by the concepts of data-driven optimization which we regard as the primary computing asset of AI.

Starting with simulation and optimization, a comprehensive overview of the different methods for different simulators is given in \cite{amaran2016simulation}. \cite{rabe2020simulation} use simulation optimization to decide on the number and location of parcel lockers to which consumer goods are delivered. \cite{munoz2013simulation} use simulation optimization to solve a complex Location Routing Problem (LRP). Applications in VRP are also abundant as shown in \cite{tripathi2009ant} for the case with stochastic demands, and \cite{perez2016simulation} with fixed time-windows. 

As for the dense combination of machine learning, mathematical optimization and simulation, there exists a few publications  - such as \cite{gutierrez2021data} - containing all three methods. \cite{rijnen2019machine} proposes an approach for urban trailer management where a machine learning model is used as a surrogate model of the simulator to evaluate simulations prematurely and reduce the computational burden while carrying out the optimization with a genetic algorithm. \cite{james2019online} employ deep reinforcement learning with neural combinatorial optimization - based on a graph neural network - to solve an online VRP. Moreover, they verify the solutions using simulation. Although the simulation is not directly involved in the optimization procedure, it is still an integral tool in solution generation.

Advancing our focus to learning and optimization where we believe most research has been done and most AI potential lies, papers like \cite{lombardi2018boosting} and \cite{bengio2021machine} survey the general integration of machine learning into combinatorial optimization for different learning mechanisms. \cite{mazyavkina2021reinforcement} survey a more detailed application of reinforcement learning to solve combinatorial optimization problems. \cite{khalil2017learning} extend on this by providing applications in graph related problems and providing a framework on how heuristics for these problems could be learned. 

Historically, the application of machine-learning to optimize VRP has been rewarding. \cite{bai2021analytics} provide a comprehensive survey of machine learning applications in solving VRPs including stochastic variants. They consider the usage of machine learning as both, a modelling tool and an optimization one. More specifically, \cite{niu2021improved} and \cite{niu2022multi} use hypothesis generation to learn a genetic algorithm to solve a multi-objective VRP with uncertain demand. For the case with stochastic customers, \cite{joe2020deep} apply reinforcement learning to approximate the value-function of actions from a genetic algorithm. 

Other applications of machine learning in solving deterministic VRP can be found in \cite{morabit2021machine} who use supervised learning for column generation to solve a VRP with time windows. \cite{furian2021machine} combine supervised learning with a branch-and-price approach to predict the value of binary decision variables in the optimal solution of an instance, and the branching scores for fractional variables for capacitated VRP. \cite{cooray2017machine} consider a different heuristic-based approach where unspervised learning is used to tune the parameters of a genetic algorithm for energy minimizing VRP.

Reinforcement learning, as a method, has been popular in the literature as well. \cite{paulo2021learning} provide a general methodology for reinforcement learning of a meta-heuristic for standard VRP with actor-critic networks. \cite{nazari2018reinforcement} employ reinforcement learning to devise a parameterized stochastic policy with an actor-critic network for optimizing capacitated VRP. They also explain that their approach could be extended to other variants. Similarly, \cite{hottung2019neural} learn a stochastic policy by reinforcement learning for capacitated VRP and split delivery VRP through an actor-critic model. However, they employ a novel concept where the training targets are defined by the objective of an infeasible solution so as to bridge the gap with the (best) feasible solution.
\cite{zhao2020hybrid} also use reinforcement learning with an actor-critic network to devise a stochastic policy whose output can be combined with a local search procedure to optimize standard VRP and VRP with time windows. 

\begin{table}
    \centering
    \resizebox{\textwidth}{!}{\begin{tabular}{c|c|c}
    \hline
    Paper& Problem& Methods\\
    \hline
    \cite{james2019online} & green logistic system online routing& reinforcement learning, combinatorial optimization\\
    \hline
    \cite{niu2021improved}&  multi-objective stochastic VRP& hypothesis generation, genetic algorithm\\
    \hline
     \cite{niu2022multi}&multi-objective stochastic VRP& hypothesis generation, genetic algorithm\\
     \hline
     \cite{joe2020deep}& stochastic VRP& reinforcement learning, genetic algorithm\\
     \hline
     \cite{morabit2021machine} &VRP with time windows& supervised learning, branch-and-price\\
     \hline
     \cite{furian2021machine}&  capacitated VRP&  supervised learning, branch-and-price\\
     \hline
     \cite{cooray2017machine}& energy minimizing VRP & unspervised learning, genetic algorithm \\
     \hline
     \cite{paulo2021learning}&  standard VRP& reinforcement learning, local search\\
     \hline
     \cite{nazari2018reinforcement}& capacitated VRP& reinforcement learning, combinatorial optimization\\
     \hline
     \cite{hottung2019neural} &split-delivery/capacitated VRP& reinforcement learning, local search\\
     \hline
     \cite{zhao2020hybrid} & standard VRP/with time windows& reinforcement learning, local search\\ 
    \hline 
    \end{tabular}}
    \caption{Categorization of the learning and optimization methods in a sample of the VRP literature.}
    \label{VRP literature categorization}
\end{table}

The aforementioned papers compose a small sample of the numerous literature on applying machine learning to solve VRP whose purpose is illustrative. Surely enough, there are other publications that address machine learning in VRP and would be of great use to a dynamic decision making system such as a DT. A summary of the papers cited above, the VRP variants they address and the employed learning and optimization methods is given in Table \ref{VRP literature categorization}.

Given the approaches prescribed above, we can easily construct a pipeline based on the AI models employed. To illustrate that, we make use of an approach that solves a stochastic VRP variant. We consider the stochasticity aspect due to its significance in characterizing urban environments where the uncertainty is due to the complex interactions between countless entities. We refer to this problem as \textit{SVRP1}.

Let the pseudo-code for solving \textit{SVRP1} be given by Algorithm \ref{SVRP method}. It is worth mentioning that Algorithm \ref{SVRP method} and the associated pipeline are examplifications whose purpose is simply illustrative. It takes as input a set of customer locations $\mathcal{N}$ with properties such as time windows and priority levels given in set $\mathcal{C}^{\mathcal{N}}$. Furthermore, let $\mathcal{T}$ represent the travel time data repository of the DT where each observation $t_{ij}$ corresponds to a travel time between locations $i$ and $j$. The output would be a set of routes $R^{*}$ representing a feasible planning. 

Algorithm \ref{SVRP method} makes use of a heuristic with a machine learning component that constantly interacts with a mathematical optimization model such as when reinforcement learning is applied to a local search heuristic to approximate the value-function of actions; see \cite{paulo2021learning}. The problem state $\mathcal{S}$ is prescribed by the current and best obtained solutions respectively, whereas the associated actions concern $k$-echange moves.

\begin{algorithm}[ht]
	\caption{Pseudo-code to solve \textit{SVRP1}.}
\begin{algorithmic}[1]
\State Input: $\mathcal{N}$, $\mathcal{C}^{\mathcal{N}}$, $\mathcal{T}$.
\State Let $\overline{R}=\emptyset$. 
\State Determine distribution of travel times $t_{ij}$ through preprocessing of $\mathcal{T}$ using statistical analysis tools.
\State Set up stochastic optimization model \textit{SVRP1}
\While {termination criterion for optimization is not met}
\State Apply search heuristic $\mathcal{H}$ that has a machine learning component.
\State Generate candidate solutions $R$ and add them to $\overline{R}$. 
\EndWhile
\For {solution $R$ in $\overline{R}$}
\State verify $R$ using simulation
\If {simulation objective of $R$ is better than $R^*$}
\State $R^*=R$.
\EndIf
\EndFor\\
\Return $R^*$
  \end{algorithmic}\label{SVRP method}
\end{algorithm}

\begin{figure}
\centering
        \includegraphics[width=\linewidth]{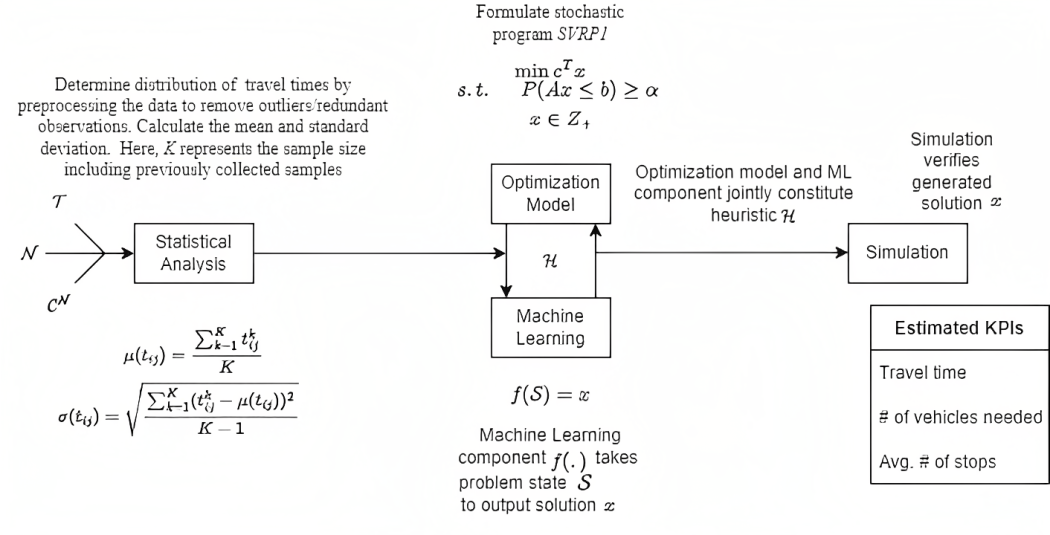}
    \caption{Example pipeline for solving \textit{SVRP1}.}
    \label{SVRP pipeline}
\end{figure}

With algorithm \ref{SVRP method}, the pipeline in Figure \ref{SVRP pipeline} can be constructed to solve \textit{SVRP1}. The pipeline illustrates the order in which entities from the AI component in figure \ref{AI component} interact with one another to output a solution $x=R$. The ontology prescribed in Figure \ref{AI component} provides a fundamental design for pipelines using the AI component. Since all relations among all entities in Figure \ref{AI component} are bidirectional, any possible ordering of the entities in a pipeline is allowed. Furthermore, other pipelines using different approaches may omit some of the entities if they do not use them at all. 

\subsection{DT Integration with AI Methods}
\label{DT Integ}

The pipeline prescribed by Figure \ref{SVRP pipeline} provide several important implications. Firstly, it is primarily dependent on the input parameters $\mathcal{T}$, $\mathcal{N}$ and $\mathcal{C^N}$. That means that as soon as there is a significant change in any of these parameters, the output of the pipeline - which is a planning verified by simulation - may very likely cease to be relevant from an optimality perspective. Secondly, it makes use of previous knowledge as specified by previously collected data on the travel times $t_{ij}$ and the machine learning component which has been subjected to training using a data-set linking operational decisions with KPIs. Lastly, it outputs a feasible operational planning.

\subsubsection{Model Specification}
\label{model specification}

With regards to the first observation, the real-time connectivity of the system is crucial. Data necessary for real-time decision-making can be collected by the DT. To illustrate this, we present the following terms. Let $P(t)$ be a high-dimensional vector representing the parameters of the underlying quantitative models of the DT at continuous time $t$. In that case, the model can be completely prescribed by $\{P(t),t>0\}$ at any arbitrary time $t$. Furthermore, let $\Delta$ be some time interval after $t$ so that $\mathcal{D}(\Delta)$ represents the data collected from the physical system in that interval. The virtual model evolution with the physical system can be characterized by the following relationship:
\begin{equation}
    P(t+\Delta)=f(\mathcal{D}(\Delta),P(t))
\end{equation}
where $f(.)$ is a concatenation of vectors of functions responsible for updating the virtual model's parameters given some data intake and previous parameter estimates. For instance, $f(.)$ could embody the series of machine learning algorithms in the DT. We elaborate more on this below. 

The specification of the functions in $f(.)$ along with the interval $\Delta$ defines the control duty of the human expert in the DT where `better' choices lead to better designs and better models. It is crucial to have a high data-updating frequency while taking latency into account as \cite{marcucci2020digital} suggest. We emphasize more on the specifications of the function $f(.)$ in Section \ref{DDO}.

\subsubsection{Data-Driven Optimization}
\label{DDO}

Here, we discuss the integration of DTs into AI pipelines to support decision-making for urban logistics. In particular, we refer to the pipeline in Figure \ref{SVRP pipeline} and the model specification in Section \ref{model specification}. For the technical aspect of integrating DTs with AI, we are not acquainted with many papers that address solving a particular-problem in urban logistics with the help of a DT except for \cite{gutierrez2021data} that deals with VRP on a high level. Other papers such as \cite{xu2022dynamic} address the detailed integration of DTs in Data-Driven Optimization (DDO), including the technical aspect, although for another application in crane-scheduling. Unfortunately, the methods do not extend naturally to other applications due to the complex structure of DTs, the physical systems they are coupled with and the relationships between them.

For our framework, we pay particular attention to the theory of machine learning and optimization as prescribed by the concepts of DDO to explain the integration process from a mathematical standpoint. Before we proceed, we propose an important categorization of DDO in a DT. We refer to DDO applications whose purpose is to amplify the virtual model's approximation of the physical system as \textbf{descriptive}. As for applications responsible for decision-making, we refer to those as \textbf{prescriptive}.

Starting with descriptive applications, the objective here is to mimic the physical system as much as possible. It is important to have an accurate representation of the physical system to guarantee correct data input when making decisions. In the example of \textit{SVRP1}, having accurate estimates of the travel times between two points is important. Otherwise, certain routes may seem favorable by the optimization mechanism, while in reality they jeopardize operational efficiency. If there is no standardized measure of the travel times, that means they have to be learned and estimated by means of some predictive model in the DT.

The utilization of predictive models to estimate variables in the physical system to provide a better representation in the virtual space has been discussed in papers like \cite{wang2020kalibre}. There, the authors discuss how a supervised learning mechanism can be deployed in the context of Bayesian optimization to estimate air-flow rates in the DT of a data center. Although the application is quite distinct from urban logistics, the methodology is quiet suitable for the learning infrastructure of DTs given their reliance on surrogate models as \cite{barkanyi2021modelling} suggest and data storage capacity as \cite{belfadel2021towards} explain. 

In the general case, learning to estimate a continuous variable - such as travel times - is a regression problem that would be guided by some loss function of the form:
\begin{equation}
\label{loss function}
    \sum_{n=1}^N\mathcal{L}(\hat{V}(x_n,\tilde{w})-V_n)
\end{equation}
aggregated over $N$ data-points. Here, $\hat{V(x_n,\tilde{w})}$ represents the output of the predictive model for the input features $x_n$ corresponding to the $n^{th}$ observation and learnable parameters $\tilde{w}$. Observe that $\tilde{w} \in P(t)$. Lastly, $V_n$ is the actual target value for the $n^{th}$ observation. The function $\mathcal{L}(.)$ could represent some deviation criterion like the square function. 

For the DT, the $N$ data points would be recalled from the  storage system where all the data that has been collected in intervals $\Delta$ - as specified in Section \ref{model specification} - is being stored. The parameters $\tilde{w}$ that characterize the predictive algorithm are optimized such that:
\begin{equation}
\label{w star}
    w^*=  \argmin_{\tilde{w}} \sum_{n=1}^N\mathcal{L}(\hat{V}(x_n,\tilde{w})-V_n) 
\end{equation}
The procedure by which the parameters $\tilde{w}$ are optimized depends on the predictive algorithm used. For instance, in the case of a neural network, an iterative procedure is used whereby $\tilde{w}$ are updated by means of a function $f_{1d}(.) \in f(.)$ that could assume a specification as follows:
\begin{equation}
\label{new w}
    \tilde{w}^{(m+1)}=f_{1d}(\tilde{w}^{(m)})=\tilde{w}^{(m)}+\alpha \cdot \hat{g}(\tilde{w}^{(m)})
\end{equation}
with $\alpha$ being a scaling parameter and $\hat{g}(\tilde{w}^{(m)})$ being an estimate of the gradient of (\ref{loss function}) at $\tilde{w}^{(m)}$; the estimate of $\tilde{w}$ at iteration $m$. The aforementioned specification of the update function $f_{1d}(.)$ is known as gradient descent and is one of many (gradient-based) optimization techniques. 

To realize how the DT's real-time data connectivity aids with training, we resort to the theory of gradient estimation for neural networks. From $N$ data-points, a batch of size $B$ is normally used to estimate $\hat{g}(\tilde{w})$ such as in:
\begin{equation}
\label{grad est}
    \hat{g}(\tilde{w})=\frac{\sum_{b=1}^B\hat{g}(x_b,\tilde{w})}{B}
\end{equation}
so that $\hat{g}(\tilde{w})$ can be seen as an average of estimates of the gradient at data points $x_b$ with parameters $\tilde{w}$ represented by $\hat{g}(x_b,\tilde{w})$. Standard training procedures select $B$ random sample from the $N$ samples in the data-set. However, if the algorithm is to be retrained in such a way so as to incorporate recent structural changes (if any), we might be tempted to estimate $\hat{g}(.)$ using $B$ samples from the most recently collected data in $\mathcal{D}(\Delta)$ during the last interval $\Delta$ since the model was updated. Previous knowledge would be incorporated in the values of $\tilde{w}^{(m)}$, which is in turn used - alongside $\mathcal{D}(\Delta)$ - to determine $\tilde{w}^{(m+1)}$ which incorporates new knowledge. 

\begin{figure}
\centering
    \includegraphics[width=\linewidth]{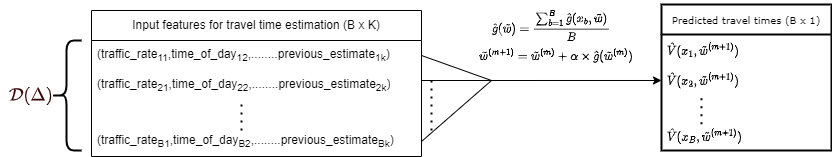}
    \caption{Abstract depiction of the estimation of $\hat{g}(.)$ using a batch $B$ from data collected.}
    \label{training with delta}
\end{figure}

Figure \ref{training with delta} abstractly portrays this training process. Here, we assume a predictive model for the travel times in a city that is based on factors like current traffic, time of the day and previous travel time estimates - assuming a time-series correlation - among other factors. 

The predictive model need not to be retrained at fixed intervals of data collection, but if the controllers believe that new data embed significant structural change in the relationship between variables, knowledge about this change ought to be incorporated in the virtual model. Structural changes in relationships between variables are referred to as ``concept drifts". \cite{lu2018learning} investigate learning under concepts drifts. With supervised learning, training is generally done offline before the predictive algorithm can be deployed by the DT to estimate the necessary parameters used as input for the formulation of mathematical optimization problems. This suffices for descriptive DDO applications.

For prescriptive ones, supervised learning generally works provided some consistency is guaranteed. For instance, if we are faced with a similar set of operational conditions, decision alternatives and KPI valuations, we could use experience gained from supervised learning to make decisions for current operations. In reality, this is seldom the case. And while consistency in operational conditions is unlikely in complex environments like cities, the decision alternatives are numerous and corresponding KPIs are unclear at first hand. 

In the context of \textit{SVRP1}, there are many possible routing options as dictated by the ordering of customers. Furthermore, it is not clear what the resulting KPIs are from each routing decision due to the associated stochasticity. In such cases, live interaction with the environment is needed to evaluate decisions produced by a local search algorithm such as k-exchange and observe the resulting KPIs. However, with logistic operations the high-cost sensitivity forbids the required trial-and-error procedure induced by reinforcement learning to generate a training data-set. To counter this issue, \cite{xu2022dynamic} propose what is currently known as offline reinforcement learning as an alternative. There, interaction with the environment occurs offline through simulation logic and standard reinforcement learning techniques could be applied again to train the algorithm before it could be deployed live for operational decision-making. The configuration for the simulation environment is determined by $P(t)$ that embeds a live representation of the status of the physical system. That said, the concepts presented by equations (\ref{loss function}), (\ref{w star}), (\ref{new w}), (\ref{grad est}) and Figure \ref{training with delta} extend themselves naturally to the case with reinforcement learning. The distinction lies in the input representation and relation to succeeding inputs, while the output evaluates the routing decision - through a reward - based on the KPIs generated from the simulation environment. The frequency of training is also important as reinforcement learning requires (re-)training after a fixed number of actions/steps has been taken, while in supervised learning, the training requirement is invariant to the number of actions taken as actions are often only produced once training has concluded.

\section{Future Research}
\label{Adv & Lim}
In analogy with the points discussed above, there are many possible benefits and challenges related to DTs. The most obvious benefit is its provision of a methodology to optimize the logistic network in a city. This would be reflected in reduced pollution and congestion volumes, more efficient logistics operations and increased consumer satisfaction through higher service levels. However, there are many costs that ought to be borne beforehand. 

For a start, \cite{botin2022digital} cite data security concerns and communication network-related obstacles. Additionally, the set-up costs of the technology-intensive DT are not negligible.
\cite{schislyaeva2021innovations} also explain that the cost-sensitivity of logistic operations may explain the reluctance of some companies to invest in testing DTs. Many LSPs may be unwilling to enable the DT to control their resources due to cost and safety concerns. The absence of a link by which the virtual model can control the physical model for testing purposes poses a serious challenge to the credibility of current studies on DTs.

To counter this, some platforms already provide basic implementations based on expert knowledge and collaboration with industry. The Atlas Leefbare Stad DT by Logistics Community Brabant \cite{ATLAS-LCB}, which is a virtual replica of cities in the Netherlands, is one such example. It models relevant variables as dictated by the requirements of its user base of academic researchers and LSPs. In Atlas, however, the transfer of data is unidirectional – from the physical to the virtual system only, in contrast to definition from Section \ref{DT definition}. \cite{marcucci2020digital} refer to such a virtual model as a Digital Shadow (DS). 

While a fully comprehensive study on DTs could not be met with a DS, a partial study is still feasible. \cite{marcucci2020digital} mention that “the primary function a DT addresses is descriptive in nature”. Examples of its descriptive functions include anomaly detection, warnings, predictive tasks and even recommending optimization-derived solutions without implementing them. By comparing its descriptive output with actual outcomes as interpreted by expert knowledge, experts can form opinions about the usefulness of the virtual model.

There are other challenges associated with building a DT. \cite{marcucci2020digital} mentions ``that technological changes and strong attention towards global warming" may require more ``radical changes in technology and policy" than the incremental approach guiding the design and development of DTs. This places pressure on the benchmarks the DT is expected to meet as the correctness of a fully functional DT may be too slow to realize any convincing gains in the short-term. 

Furthermore, \cite{marcucci2020digital} explain that relationships between variables is expected to change over the course of time due to external factors . The DT model, therefore, compels constant updates so that changes in relationships and knowledge are incorporated on time, otherwise its added value may be questionable. 

On the other hand, with past data, DTs can explain the possible underlying causes of encountered phenomena. This real-time management and control of situations aids with the integration of short-term decision making with long-term strategies as \cite{nochta2019governance} suggests. Therefore, the DT would provide a more suitable framework to achieve the sustainability goals than other contemporary methodologies. That said, we aspire that future research expands on the ontology we proposed.

\section*{Acknowledgements}
Abdo Abouelrous is supported by the AI Planner of the Future programme, which is supported by the European Supply Chain Forum (ESCF), The Eindhoven Artificial Intelligence Systems Institute (EAISI), the Logistics Community Brabant (LCB) and the Department of Industrial Engineering and Innovation Sciences (IE\&IS).

\bibliographystyle{model5-names}
\bibliography{MyBIB}
\end{document}